\documentclass[10pt,twocolumn,letterpaper]{article}

\usepackage[final]{cvpr}      

\usepackage{graphicx}
\usepackage{amsmath}
\usepackage{amssymb}
\usepackage{booktabs}

\usepackage[pagebackref,breaklinks,colorlinks]{hyperref}

\usepackage[capitalize]{cleveref}
\crefname{section}{Sec.}{Secs.}
\Crefname{section}{Section}{Sections}
\Crefname{table}{Table}{Tables}
\crefname{table}{Tab.}{Tabs.}

\usepackage{graphicx}

\usepackage{xcolor}

\def\cvprPaperID{*****} 
\def\confName{CVPR}
\def\confYear{2022}

\begin{document}

\title{Importance is in your attention: \\ agent importance prediction for autonomous driving}

\author{Christopher Hazard, Akshay Bhagat, Balarama Raju Buddharaju, Zhongtao Liu, \\
Yunming Shao, Lu Lu, Sammy Omari, Henggang Cui \\
Motional \\
}
\maketitle

\begin{abstract}
Trajectory prediction is an important task in autonomous driving.
State-of-the-art trajectory prediction models often use attention mechanisms to model
the interaction between agents.
In this paper, we show that the attention information from such models can also be used to
measure the importance of each agent with respect to the ego vehicle's future planned trajectory.
Our experiment results on the nuPlans dataset show that
our method can effectively find and rank surrounding agents by their impact on the ego's plan.
\end{abstract}

\setlength{\belowdisplayskip}{0pt} \setlength{\belowdisplayshortskip}{0pt}
\setlength{\abovedisplayskip}{0pt} \setlength{\abovedisplayshortskip}{0pt}

\section{Introduction}

In order to navigate in the dynamic environment, the autonomous vehicle needs to
detect the current locations of the other agents in the environment and predict their future trajectories.
Start-of-the-art trajectory prediction models use deep neural networks with attention mechanisms~\cite{transformer, gat} to
model the interactions between agents~\cite{lanegcn, gohome, thomas, vectornet, scenetransformer, mmtransformer}.
Those prediction models also often include the ego vehicle in the interaction graph
in order to model the interactions between the other agents and the ego vehicle.

In addition to the predicted future trajectories, a downstream motion planning module consuming these predictions
can also benefit from knowing
how much another agent is likely to affect the future maneuvering of the ego vehicle~\cite{refaat2019agent}.
For example, an agent that is currently behind and traveling away from the ego
is not likely to have much impact on the ego's plan while a vehicle making a lane change in front of the ego is very significant.
With this knowledge, the motion planner module can focus its computational resources
on handling the more important agents and potentially use coarser-grained methods to handle agents with low importance.

The focus of this work is to predict the importance score of the other agents.
The most straightforward approach is to model the importance prediction task as a classification problem and
train a prediction head using supervised training.
However, this approach requires the ground-truth labels
for the agent importance scores, and such labels can be hard to obtain.

In this paper, we propose a simple method to predict the importance score of the other agents
without requiring any extra training labels that takes advantage of the fact that most of the 
state-of-the-art trajectory prediction models already use a built-in attention mechanism to model the interactions between agents in a graph.
Through a series of experiments, we show that the attention weights between the ego vehicle and other agents
can naturally represent the degree to which the existence of each other agent affects the predicted maneuvering of the ego vehicle.

\section{Related work}
Agent importance prediction is often used in autonomous driving stacks~\cite{refaat2019agent},
typically calculated with the use of human labeled ground-truth importance scores.
Not only are these expensive to label, but label quality is hard to control due to the subjective nature of this task.
To address this limitation, \cite{refaat2019agent} proposes to generate the ground-truth labels
by running an existing planner in simulation on the dataset and labeling the agent importance based on the planner cost.
Our method, on the other hand, does not require any ground-truth labels to train.

The trajectory prediction task predicts the future trajectories of a set of agents,
given their history tracks and map information.
Since the behavior of an agent also depends on the state of the other agents,
it is important for the trajectory prediction model to be able to model the interactions between agents
when making predictions.
The graph attention mechanism~\cite{transformer, gat} is the most popular approach for modeling such agent interactions.

LaneGCN~\cite{lanegcn} proposes an Agent-to-Agent attention module to model the agent interactions,
and it is also later used by GOHOME~\cite{gohome} and THOMAS~\cite{thomas}.
Given the agent input features $\{x_i\}_{i=1}^{N}$,
the Agent-to-Agent attention module computes the agent output features $\{y_i\}_{i=1}^{N}$ as:

\setlength{\belowdisplayskip}{0pt} \setlength{\belowdisplayshortskip}{0pt}
\setlength{\abovedisplayskip}{0pt} \setlength{\abovedisplayshortskip}{0pt}

\begin{align}
\label{eq:lanegcn_attention}
y_i = x_i W_0 + \sum_j \phi(\text{concat}(x_i, \Delta_{i,j}, x_j) W_1) W_2
\end{align}

VectorNet~\cite{vectornet}, SceneTransformer~\cite{scenetransformer}, and mmTransformer~\cite{mmtransformer}
uses the Transformer attention module proposed in~\cite{transformer}.
Given the agent input features $\{x_i\}_{i=1}^{N}$, Transformer attention
computes the agent output features query $\{y_i\}_{i=1}^{N}$
using $Q$, key $K$, and value $V$ matrices:

\begin{align}
\label{eq:transformer_attention}
{\bf y} = \text{softmax}(\frac{Q({\bf x}) K({\bf x})^T}{\sqrt{d}}) V({\bf x})
\end{align}

In order to model the interactions between the other agents and the ego vehicle,
those prediction models also often include the ego vehicle in the interaction graph
and predict the future trajectory of the ego vehicle in the same way as the other agents.

\section{Agent importance prediction}

\subsection{Problem setup}

We propose computing the importance score of the agents by
using their attention weights with respect to the ego vehicle from the agent to agent interaction module of the trajectory prediction model.
The main inputs of this module are the feature vectors of all $N$ actors in the scene,
denoted as $\{x_i\}_{i=1}^{N}$.
The outputs of this module are the output feature vectors with the actor interactions modeled,
denoted as $\{y_i\}_{i=1}^{N}$.

\setlength{\belowdisplayskip}{0pt} \setlength{\belowdisplayshortskip}{0pt}

\begin{align}
y_i = \text{Interaction}(\{x_j\}_{j=1}^{N})
\end{align}

The goal of our work is to predict the importance scores $\gamma$ from an agent (with feature $x_a$)
to the ego vehicle (with feature $x_e$)
using a pretrained \texttt{Interaction} module.

\subsection{Attention in a single attention}

When there is a single attention layer in the \texttt{Interaction} module,
it usually has the property that
the contributions from each agent $j$ to agent $i$ are computed with a function $g$ and then summed together.

\begin{align}
y_i = f(x_i) + \frac {\sum_j g(x_i, x_j)}{\text{Normalizer}}
\end{align}

With this formulation, we propose to define $g(x_i, x_j)$ as the \emph{attention vector}
from agent $j$ to agent $i$,
and use its L2 norm as the importance score from agent $j$ to agent $i$,
which represents the magnitude of agent $j$'s influence on the future trajectory predictions of agent $i$.

\begin{align}
\gamma(x_i, x_j) = \| g(x_i, x_j) \|_2
\end{align}

This formulation generalizes the attention modules used by most of the state-of-the-art trajectory prediction works,
including the LaneGCN attention module
used in LaneGCN~\cite{lanegcn}, GOHOME~\cite{gohome}, and THOMAS~\cite{thomas},
and the Transformer attention module used in VectorNet and SceneTransformer.

For LaneGCN attention (Eq~\ref{eq:lanegcn_attention}), the attention vector is simply

\begin{align}
\label{eq:lanegcn_attention_vector}
g(x_i, x_j) = \phi(\text{concat}(x_i, \Delta_{i,j}, x_j) W_1) W_2
\end{align}

The Transformer attention (Eq~\ref{eq:transformer_attention}) has the softmax operation,
but we can expand its formula as

\begin{align}
\label{eq:transformer_attention_vector}
\begin{split}
y_i &= \sum_j \frac{e^{Q(x_i) K(x_j)^T}}{\sqrt{d} \sum_k e^{Q(x_i) K(x_k)^T}} V(x_j) \\
    &= \frac{\sum_j e^{Q(x_i) K(x_j)^T)} V(x_j)}{\sqrt{d} \sum_j e^{Q(x_i) K(x_j)^T}}
\end{split}
\end{align}

Which gives us

\begin{align}
\label{eq:transformer_attention_vector}
g(x_i, x_j) = e^{Q(x_i) K(x_j)^T)} V(x_j)
\end{align}

\subsection{Attention in multiple layers}

When there are multiple attention layers in the \texttt{Interaction} module,
we can compute the agent importance scores on each of the attention layers and aggregate them.
Our evaluation result in Section~\ref{sec:eval_multiple_layers} shows that
we get similar performance by taking the average importance score, maximum importance score,
or just the importance score from the last attention layer.

\section{Evaluation}

\subsection{Experiment setup}

We built our agent importance prediction module on top of the LaneGCN~\cite{lanegcn} model,
which shares the same Agent-to-Agent attention module as GOHOME~\cite{gohome} and THOMAS~\cite{thomas}.
We trained the model on the nuPlan~\cite{nuplan} training dataset
and ran the trained model on 2000 randomly selected validation samples.
The prediction horizon is 8 seconds.
The LaneGCN model contains two attention layers for Agent-to-Agent attention,
and by default, we computed the agent importance scores from the last attention layer.
Since we use agent attention value to compute the agent importance score,
we will use ``attention value'' and ``importance score'' interchangeably in this section.



\subsection{Correlation between agent attention and ego behavior change}

In this set of experiments, we show that
the agents with high predicted importance scores
are indeed the ones that have high impacts on the ego behavior.
To demonstrate this, we sort the agents in each scene by their importance score,
remove each of them, and measure how the predicted ego trajectory will change.
\footnote{
    Here we use the ego trajectory predicted from the prediction model as a proxy to
    the ego plan from the motion planner.
}

We show the results in Table~\ref{tab:agent_removal}.
We calculate the correlation between the normalized attention value (i.e., importance score) assigned to the removed agent and the pointwise L2 distance of the predicted ego trajectory
before and after removing the agent, as well as the correlation to the amount of change in the prediction L2 error (w.r.t. ground-truth ego trajectory) before and after removal.
We report both Pearson correlation and R-squared values for each quantity.
The R-squared values correspond to the amount of variance explained by the dependent variable in a linear model.

\begin{table}[h!]
    \footnotesize
    \begin{center}
    \begin{tabular}{||c | c c | c c|| l p{.25cm}} 
        \hline
         & \multicolumn{2}{c|}{\bf Pred traj delta corr.} & \multicolumn{2}{c||}{\bf Pred error delta corr.} \\
        {\bf $k$-th Agent} & {\bf Pearson} & {\bf $R^2$} & {\bf Pearson} & {\bf $R^2$} \\
        \hline\hline
        1 & .477 & .128 & .228  & .016\\
        \hline
        2 & .341 & .073 & .116  & .005\\
        \hline
        3 & .211 & .028 & .044  & .0008\\
        \hline
        All & .200 & .117 & .040 & .014\\
        \hline
    \end{tabular}
    \caption{
        We remove the $k$-th highest attended agent,
        and show the correlation between the predicted ego trajectory delta and attention value,
        as well as the correlation between the predicted ego trajectory error (w.r.t. ground-truth ego trajectory) delta and attention value.
        We report both Pearson correlation and R-squared values for each quantity.
        The last row contains the results for an experiment in which we remove all other agents in the scene,
        and we compute the correlation using the sum of the attention values of all agents.
    }
    \label{tab:agent_removal}
    \end{center}
\end{table}

The correlation between the attention value of the agent removed and the change in predicted ego trajectory is highly positive,
indicating that our method indeed assigns high importance scores to agents that have high impacts on the ego plan.
We also observe that the correlation decreases as $k$ gets bigger,
meaning the ego trajectory is more correlated with the higher attended agents.

The last row of Table~\ref{tab:agent_removal}, in which we remove all other attended agents, shows a reduction in the Pearson
correlation with regards to each of the other rows despite the fact that this experiment removes the most attention.
The fact that the Pearson correlation is so small indicates that overall the 
total removed attention was less effective than the same amount of attention attributed to the other single removal experiments. 
Given that the Pearson correlations decrease with regards to rank, we can conclude that
our model is rightfully assigning more attention to the most important agents as desired.

\begin{figure}
    \centering
\begin{subfigure}{.25\textwidth}
    \includegraphics[width=4cm]{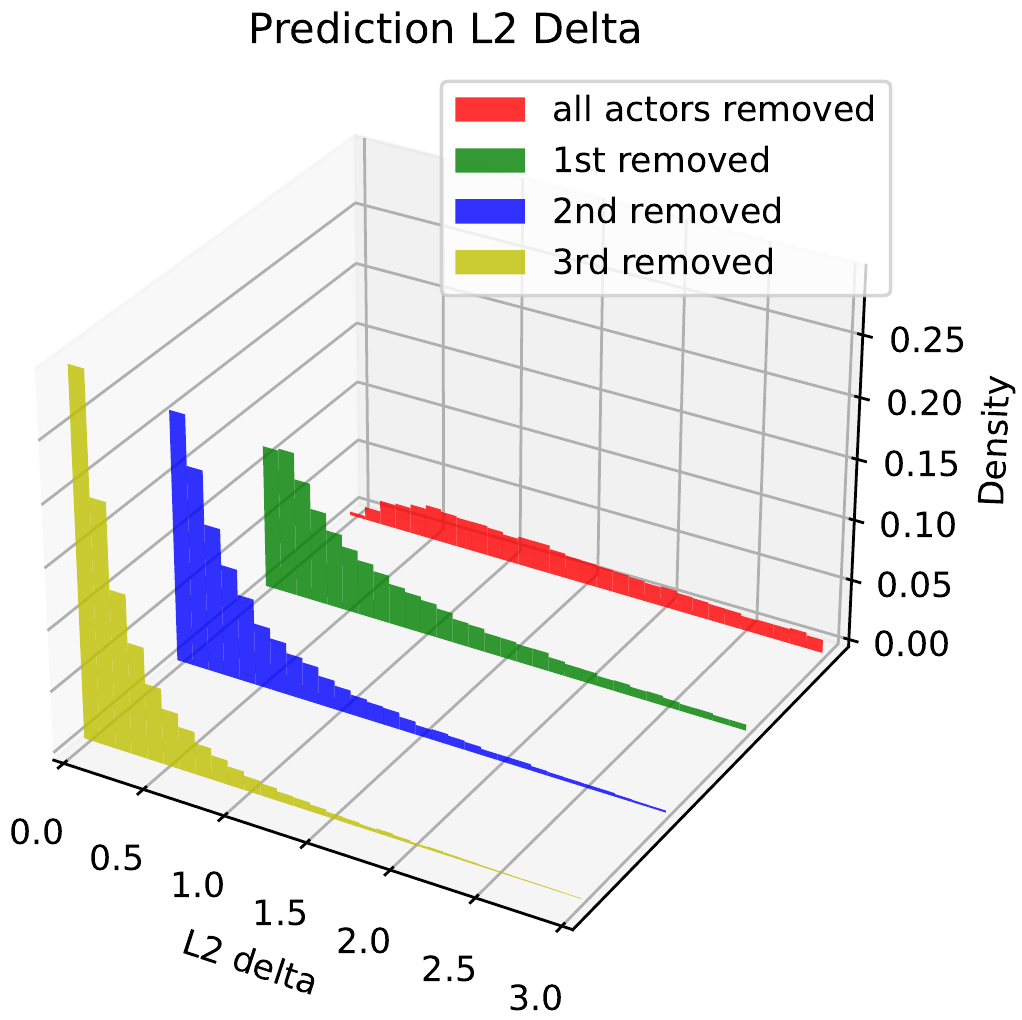}
    \label{fig:sub1}
\end{subfigure}
 \begin{subfigure}{.2\textwidth}
    \includegraphics[width=4cm]{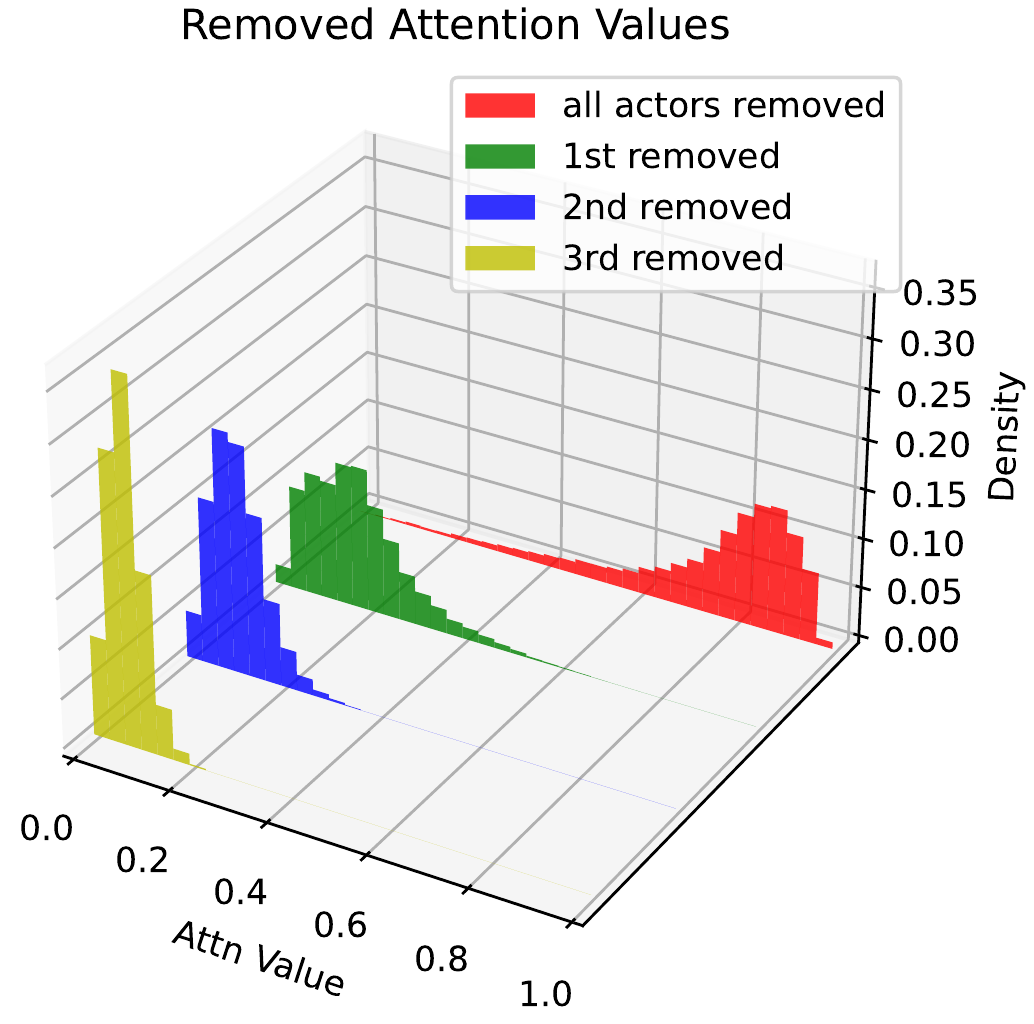}
    \label{fig:sub2}
 \end{subfigure}
\caption{
    Distribution of ego trajectory L2 delta between pre- and post- removal of the highest attended agent and the amount of attention removed in each experiment.
    We note that the plot of L2 distances has its domain scaled between (0,3) meters to show the bulk of the density, however there are a significant number
    of outliers going out to as far as 20 meters in total error, meaning that our distributions have very long tails.
} 
\label{fig:distribution}
\end{figure}

Figure~\ref{fig:distribution} shows the distribution of ego trajectory delta between pre- and post- removal
of the highest attended agent.
It also confirms the conclusion that agents that are predicted to have high importance scores (i.e., high attention values)
also have high importance on the ego behavior.

\begin{table}[h!]
    \footnotesize
    \begin{center}
    \begin{tabular}{||c c c c c c c|| l p{.8cm} }
        \hline
        {\bf Quantity} & {\bf 0} & {\bf 30} & {\bf 50} & {\bf 80} & {\bf 90} & {\bf 100} \\
        \hline\hline
        Highest attns & .003 & .117 & .165 & .245 & .307 & .748 \\
        \hline
        \# relevent agents & 0 & 1 & 2 & 3 & 3 & 7\\
        \hline
        Max attn shift & .003 & .306 & .371 & .485 & .553 & .812 \\
        \hline
        Traj anglar delta [rad] & 0 & .001 & .002 & .01 & .044 & .283 \\
        \hline
    \end{tabular}
    \caption{
        Quantile distributions.
    }
    \label{tab:quantiles}
    \end{center}
\end{table}

In Table~\ref{tab:quantiles}, we show the quantile distribution of
1) attention value of the highest attended agent,
2) the number of agents with at least 0.1 normalized attention value,
3) maximum (among all agents) attention changes between the two attention layers,
and 4) trajectory angular delta.
The trajectory angle is defined as the angle of the vector from the ego's current position to the last predicted waypoint, and we compute the delta between the pre- and post- agent removal predictions.
We find the vast majority of ego vehicles experience very small and typically negligible maximum angle change.
However, in some rare cases, there are significant angular deltas, 
such as when the highest likelihood modality changes at an upcoming fork in the road depending on the agent in front of the ego,
or more commonly when our prediction switches to an adjacent lane change.

\subsection{Spatial attention distribution}
\begin{figure}
    \centering
    \begin{subfigure}{1.0\linewidth}
        \includegraphics[width=.9\linewidth]{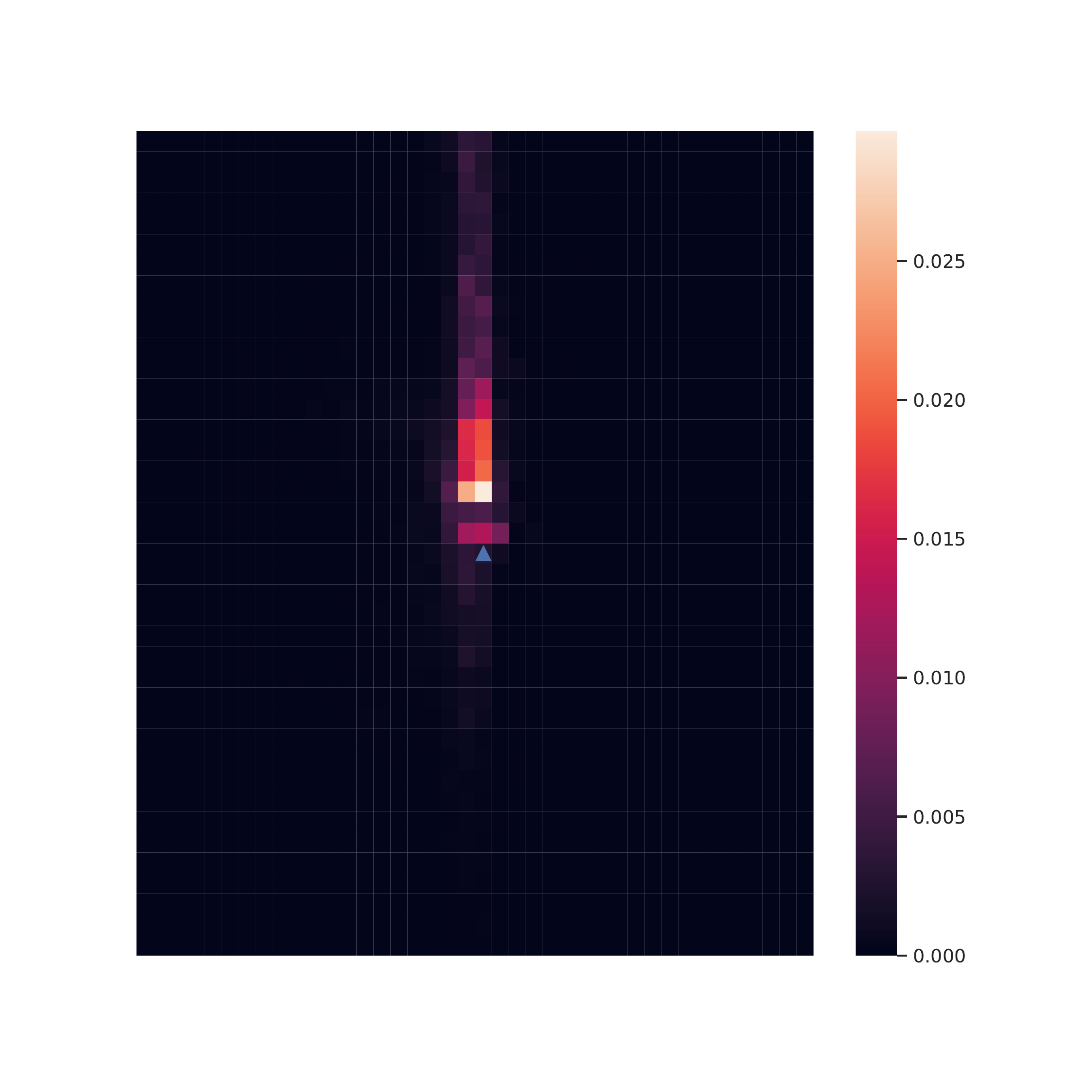}
    \end{subfigure}
    \caption{
        Normalized attention per $4 \times 4$ meter block in the ego-centric coordinate frame.
        The ego position and orientation is marked with the blue triangle (in the middle and pointing north).
    }
    \label{fig:attention_heatmap}
\end{figure}

In Figure~\ref{fig:attention_heatmap}, we plot the average normalized attention value (i.e. importance score)
per $4 \times 4$ meter block in the ego-centric coordinate frame.
We can see that the majority of the attention is placed on agents
directly in front of our ego agent since those are the agents whose future behavior would most likely cause the ego agent to behave differently.
In comparison, very little attention is typically paid to vehicles behind the ego since their presence usually wouldn't impact how the ego vehicle should behave unless
the ego needed to adjust its path to avoid a future collision with the agent behind it.

\subsection{Attention in multiple layers}
\label{sec:eval_multiple_layers}

\begin{table}[h!]
\footnotesize
\begin{center}
\begin{tabular}{||c | c c|| l p{.35cm} } 
    \hline
    & \multicolumn{2}{c||}{\bf Pred traj delta corr.} \\
    {\bf Aggre Function} & {\bf Pearson} & {\bf $R^2$} \\
    \hline\hline
    Max & .524 & .275\\
    \hline
    Mean & .498 & .248\\
    \hline
    Last & .512 & .262 \\
    \hline
\end{tabular}
\caption{Correlation between ego trajectory delta and attention values with different attention aggregation functions.}
\label{tab:multiple_layers}
\end{center}
\end{table}

In this experiment, we perform an ablation study to compare different aggregation functions
to aggregate the attention values (i.e., importance scores) from multiple attention layers.
In Table~\ref{tab:multiple_layers}, we compared using three aggregation functions:
1) maximum attention among all layers, 2) average attention among all layers,
and 3) only use the attention of the last layer.
We compared the correlation between ego trajectory delta and the importance score,
and the result shows that all three functions yield similar results,
which means our method is robust to the selection of aggregation function.

\section{Conclusion and future works}

In this work, we have studied the allocation of attention to agents surrounding our ego vehicle and shown that
the normalized magnitude of the attention vector produced by the model for each agent is a good indicator of 
the underlying importance of each agent.
By using those attention values as the agent importance scores,
we are able to properly prioritize agents that will have high impacts on the ego trajectory.

In our evaluation we used the ego trajectory predicted from the prediction model as a proxy for
the ego plan from a motion planner: in future work we will integrate this importance prediction module into a 
real motion planner.

{\small
\bibliographystyle{ieee_fullname}
\bibliography{main}
}

\end{document}